\documentclass[runningheads]{llncs}

\usepackage{multirow} 
\usepackage{array}     
\usepackage{booktabs}
\usepackage[T1]{fontenc}
\usepackage[normalem]{ulem}
\useunder{\uline}{\ul}{}
\usepackage{graphicx,verbatim}
\usepackage{graphicx}
\usepackage{amsfonts,amssymb}
\usepackage{caption}
\usepackage{color}
\usepackage{epstopdf}
\usepackage{tabularx,threeparttable}
\usepackage{array}
\usepackage[hidelinks]{hyperref}
\usepackage{graphicx}
\usepackage{subfigure}
\usepackage{amssymb, amsmath, bm}
\usepackage{mathtools}
\usepackage{mathrsfs}
\usepackage{booktabs}
\usepackage{multirow}
\usepackage{cite}
\usepackage{url}
\usepackage{color}
\usepackage{dsfont}
\usepackage{enumitem}
\usepackage[table,xcdraw]{xcolor}
\usepackage{hyperref}
\hypersetup{
    colorlinks = true,    
    citecolor = blue      
}

\begin{document}
\title{Ada-FCN: Adaptive Frequency-Coupled Network for  fMRI-Based Brain Disorder Classification}

\titlerunning{Ada-FCN for Brain Disorder Classification}

\author{
    Yue Xun\inst{1} \and
    Jiaxing Xu\inst{2} \and
    Wenbo Gao\inst{1} \and
    Chen Yang\inst{1} \and
    Shujun Wang\inst{1,3,4}\thanks{Corresponding author.}
}
%
\index{Xun, Yue}
\index{Xu, Jiaxing}
\index{Gao, Wenbo}
\index{Yang, Chen}
\index{Wang, Shujun}
\authorrunning{Y. Xun et al.}
%
\institute{
    Department of Biomedical Engineering, Hong Kong Polytechnic University, Hong Kong SAR \\
    \email{shu-jun.wang@polyu.edu.hk} \and
    College of Computing and Data Science, Nanyang Technological University, Singapore \and
    Research Institute for Smart Ageing, The Hong Kong Polytechnic University, Hong Kong SAR \and
    Research Institute for Artificial Intelligence of Things, The Hong Kong Polytechnic University, Hong Kong SAR
}
    
\maketitle              
\begin{abstract}








Resting-state fMRI has become a valuable tool for classifying brain disorders and constructing brain functional connectivity networks by tracking BOLD signals across brain regions. 
However, existing models largely neglect the multi-frequency nature of neuronal oscillations, treating BOLD signals as monolithic time series.  This overlooks the crucial fact that neurological disorders often manifest as disruptions within specific frequency bands, limiting diagnostic sensitivity and specificity. While some methods have attempted to incorporate frequency information, they often rely on predefined frequency bands, which may not be optimal for capturing individual variability or disease-specific alterations. 
To address this, we propose a novel framework featuring Adaptive Cascade Decomposition to learn task-relevant frequency sub-bands for each brain region and Frequency-Coupled Connectivity Learning to capture both intra- and nuanced cross-band interactions in a unified functional network. 
This unified network informs a novel message-passing mechanism within our Unified-GCN, generating refined node representations for diagnostic prediction.  Experimental results on the ADNI and ABIDE datasets demonstrate superior performance over existing methods. The code is available at \url{https://github.com/XXYY20221234/Ada-FCN}.


\end{abstract}
\section{Introduction}



In the field of neuroscience, a key aim is to derive abnormal patterns in the brain that are linked to neurological disorders such as Alzheimer’s, Autism, and Parkinson’s. Resting-state state functional magnetic resonance imaging (fMRI) has been becoming a valuable technique for this objective by tracking Blood Oxygenation Level Dependent (BOLD) signals across paired brain regions and then constructing brain functional connectivity network where nodes represent different brain regions and edges indicate signal correlations between them~\cite{cui2022interpretable}. The brain functional network helps identify synchronized activity that could serve as dynamic biomarkers for neurological disorders, facilitating early diagnosis and treatment.

Recent advances in graph neural networks (GNNs) have spurred significant progress in brain functional network analysis~\cite{kan2022brain,xu2024contrastive,peng2025biologically,xu2025brainood}. Models like BrainGNN~\cite{li2021braingnn} employ ROI-selection pooling to highlight disease-relevant regions, while PRGNN~\cite{li2020pooling} enforces group-level consistency through graph pooling with anatomical regularization. Complementary approaches, such as BrainNetCNN~\cite{kawahara2017brainnetcnn}, leverage edge-to-edge convolutional filters to exploit topological relationships, and Transformer-based architectures~\cite{ying2021transformers} utilize attention mechanisms to model global interdependencies. Despite their success, these methods share a critical limitation: they treat BOLD signals as monolithic time series, disregarding the multi-frequency nature of neuronal oscillations~\cite{zuo2010oscillating}. This challenge of identifying and leveraging key frequency components is also a central topic in the broader field of multivariate time series forecasting~\cite{yu2025refocusreinforcingmidfrequencykeyfrequency,yu2025linoadvancingrecursiveresidual}.
It is worth noting that different neurological disorders often manifest as disruptions in specific frequency bands~\cite{yang2020frequency}. Therefore, the conventional practice of restricting brain functional connectivity analysis to a single, low-frequency band can obscure critical, band-specific information present in other frequencies, leading to a loss of sensitivity and specificity in disease diagnosis.

Recognizing this, some work has attempted to incorporate frequency information into brain functional network analysis. Hu \textit{et al.}~\cite{hu2021multi} used the discrete wavelet transform (DWT) to decompose BOLD signals into multiple frequency bands and then constructed a sparse functional connectivity network by fusing the information from each band. MFHC~\cite{zhang2017constructing} constructed both frequency-specific and cross-frequency high-order functional connectivity networks to capture richer interaction patterns.  Tewarie \textit{et al.}~\cite{tewarie2016integrating} used a multiplexed graph representation to analyze cross-frequency interactions in magnetoencephalography (MEG) data. While existing frequency-based methods consider multi-band information, they suffer from two critical limitations. First, they rely on preset frequency ranges despite the fact that individual brain functions may work differently, especially when they have neurological disorders. This means that important brain connectivity patterns related to the disease might be missed. Second, they simply combine features from different frequency bands, but signals from distinct frequencies represent different physiological processes. Mixing them directly may fail to reveal interactions between frequencies and lead to misleading conclusions.

To address these limitations,  we propose a novel framework that involves Adaptive Cascade Decomposer and Frequency-Coupled Connectivity Learning to enhance the analysis of brain functional connectivity. We adaptively decompose the raw fMRI time series into task-relevant frequency sub-bands for each brain region. This learned decomposition involves tailored low- and high-frequency signal pairs. We then construct a comprehensive functional connectivity network, capturing both intra-band and, importantly, nuanced cross-band interactions. These connectivity patterns are unified into a single representation of the brain’s functional network. This unified network is the basis for a novel message-passing mechanism within our Unified-GCN. The GCN generates refined node representations, which are aggregated and processed by fully connected layers to produce the final diagnostic prediction. Our contributions can be summarized as follows:
\vspace{-15pt}
\begin{itemize}[label=\textbullet]
    \item We propose a learnable decomposition method to adaptively extract task-relevant frequency sub-bands from fMRI time series, overcoming the limitations of fixed or handcrafted frequency band definitions in existing approaches.
    \item We introduce a Unified-GCN framework, incorporating a novel Dual-Projection Bilinear Attention mechanism for holistic brain network modeling. This end-to-end framework seamlessly integrates adaptive decomposition, frequency-aware message passing, and cross-frequency alignment through attention.
    \item Extensively evaluated on the ADNI and ABIDE datasets, our model demonstrates superior diagnostic accuracy and AUROC compared to state-of-the-art methods.
\end{itemize}

\section{Method}
Neurological disorder classification based on brain functional networks aims to predict the state of the disease of each subject by using fMRI signals. Given a set of fMRI time series $X \in \mathbb{R}^{N \times T}$, which represents the activity of $N$ brain ROIs in $T$ time points. Our objective is to learn a predictive function $f: X \to y \in \{1, 2, \ldots, c\}$, which maps the fMRI signals to a diagnostic label. 



\begin{figure}[!t]
\includegraphics[width=\textwidth]{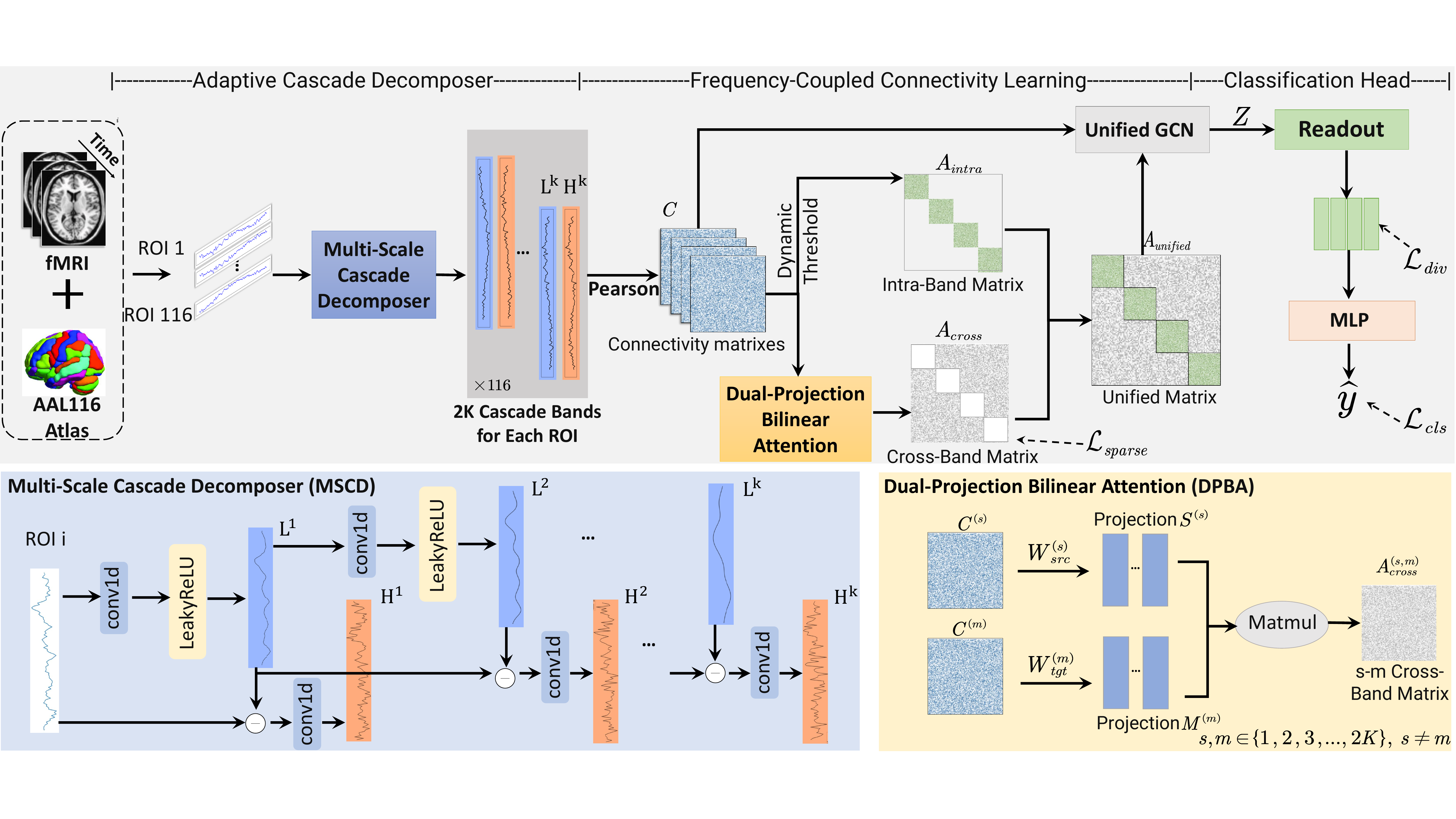}
\caption{The framework of Ada-FCN for fMRI-based brain disorder classification.} \label{pic1}
\end{figure}

As illustrated in Fig.~\ref{pic1}, Ada-FCN consists of three key components: an adaptive cascade decomposer, a frequency-coupled connectivity learning network, and classification head and loss function.

\subsection{Adaptive Cascade Decomposer}
Existing methods for fMRI-based disorder identification often overlook the complex interplay between different frequency components, and preset or full bands might not optimally capture the diverse neural information present across different tasks or individuals. To address this limitation, we introduce an adaptive cascade decomposer that adaptively decomposes the original time series into a hierarchy of frequency sub-bands. Given $X = [x_1, \ldots, x_N]^\top \in \mathbb{R}^{N \times T}$, where $\mathbf{x}_i \in \mathbb{R}^T$ denotes the time series of the $i$-th ROI, We define $L_i^0 = x_i$ as the initial input. For each ROI $i$, and at each level $k \in \{1, \ldots, K\}$, the decomposition proceeds in two steps: 
\begin{equation}
L_i^k = \mathcal{F}_L^k\bigl(L_i^{k-1}\bigr), \quad
H_i^k = \mathcal{F}_H^k\Bigl(L_i^{k-1} - L_i^k\Bigr), \tag{1}
\end{equation}
where $L_i^k$ is the low-frequency approximation at level $k$ and $H_i^k$ is the corresponding high-frequency residual. The low-frequency extraction operator $\mathcal{F}_L^k(\cdot)$ is implemented using a dilated 1D convolution with an appropriately chosen kernel, padding, and dilation factor, followed by a LeakyReLU activation. In contrast, the high-frequency extraction operator $\mathcal{F}_H^k(\cdot)$ employs a separate 1D convolution with a smaller kernel and standard padding, without an intervening activation function to refine the residual signal.  In this manner, for each ROI $i$, we obtain a cascade of sub-band signals:
\begin{equation}
\widetilde{X} = \{(L^1_i, H^1_i), \dots, (L^K_i, H^K_i)\}\,,
\tag{2}
\end{equation} where each pair $(L_i^{k}, H_i^{k})$ represents the low-frequency approximation and high-frequency residual at the $k$-th scale.
By stacking these 2$K$ sub-band signals along a new dimension, the original fMRI time series $X \in \mathbb{R}^{N \times T}$ is transformed into a multi-band representation $\widetilde{X} \in \mathbb{R}^{(2K) \times N \times T}$.

\subsection{Frequency-Coupled Connectivity Learning}
\subsubsection{Intra-Band Connectivity via Dynamic Thresholding.} 
Intra-band connectivity is crucial because each frequency band offers distinct insights into brain function. We use an adaptive dynamic threshold to retain robust connections tailored to each band’s statistics, effectively filtering out noise.
For each subject's decomposed sub-bands $\widetilde{X} \in \mathbb{R}^{(2K) \times N \times T}$, we compute the Pearson correlation to generate a connectivity matrix $C^{(k)} = \operatorname{Pearson}\Bigl(\widetilde{X}^{(k)}\Bigr) \in \mathbb{R}^{N \times N}$, $k = 1, \ldots, 2K$, then apply an dynamic threshold to eliminate weak connections:
\begin{equation}
A_{\mathrm{intra}}^{(k)} = T_{\tau k}\bigl(C^{(k)}\bigr), \quad
T_{\tau k}(\cdot): \operatorname{mask}_{ij} =
\begin{cases}
1, & \text{if } \lvert C_{ij}^{(k)} \rvert > \tau_k,\\[1mm]
0, & \text{otherwise},
\end{cases}.
\tag{3}
\end{equation}
These 2$K$ sparsified intra-band matrices are then integrated using the Kronecker direct sum $A_{\mathrm{intra}} = \bigoplus_{k=1}^{2K} A_{\mathrm{intra}}^{(k)} \in \mathbb{R}^{(2KN) \times (2KN)}$ which is a block-diagonal matrix capturing intra-band transitions within each sub-band.
\subsubsection{Cross-Band Coupling with Dual-Projection Bilinear Attention.} To model interactions across frequency bands, we adopt an attention-based coupling mechanism that learns off-diagonal adjacency blocks. 
Consider two distinct frequency bands $s, m \in \{1,2,3,\ldots,2K\}, s \neq m$. Let $C^{(s)},\; C^{(m)} \in \mathbb{R}^{N \times N}$ be their correlation matrices. First, each matrix is transformed into source or target spaces via learnable weights $W_{src}^{(s)},\; W_{tgt}^{(m)} \in \mathbb{R}^{d \times d}$, where $d$ is the hidden dimension. Concretely, we obtain $S^{(s)} = C^{(s)}\, W_{src}^{(s)} \in \mathbb{R}^{N \times d}$ and $M^{(m)} = C^{(m)}\, W_{tgt}^{(m)} \in \mathbb{R}^{N \times d}$. This dual-projection allows asymmetric and band-specific interactions. Next, we compute the bilinear interaction $S^{(s)} \bigl(M^{(m)}\bigr)^\top$, where each entry in the resulting $\mathbb{R}^{N \times N}$ matrix captures the pairwise similarity between the $i$-th row of $S^{(s)}$ and the $i$-th row of $M^{(m)}$:
\begin{equation}
A_{\text{cross}}^{(s,m)} = S^{(s)} \bigl(M^{(m)}\bigr)^\top.
\tag{4}
\end{equation}
Finally, we construct  $A_{\mathrm{cross}} \in \mathbb{R}^{(2KN) \times (2KN)}$ by arranging all cross-band matrices $\mathbf{A}_{\text{cross}}^{(s,m)}$ at off-diagonal blocks corresponding to band pairs, while setting the diagonal blocks to zero.

\subsubsection{Unified Graph Convolution.}
The unified graph convolution integrates both intra-band and cross-band connectivity patterns to enable holistic message passing across all frequency-specific nodes, by:
\begin{equation}
A_{\mathrm{unified}} = A_{\mathrm{intra}} + \lambda A_{\mathrm{cross}} \in \mathbb{R}^{(2KN) \times (2KN)},
\tag{5}
\end{equation}
\begin{equation}
H_0 = \operatorname{Stack}\Bigl( C^{(1)},\, C^{(2)},\, \ldots,\, C^{(2K)} \Bigr) \in \mathbb{R}^{(2KN) \times N},
\tag{6}
\end{equation}
where $\lambda$ is a learnable scaling factor that balances the contribution of cross-band interactions. The initial node features $H_0$ are derived from the raw Pearson correlation matrices. These are used for unified graph convolution:
\begin{equation}
H^{(i)} = \sigma\Bigl( D^{-\frac{1}{2}} A_{\text{unified}} D^{-\frac{1}{2}} H^{(i-1)} W^{(i)} \Bigr),
\tag{7}
\end{equation}
where $W^{(i)}$ is the learnable weight matrix for the $i$-th layer, D is the degree matrix computed from the unified adjacency matrix $A_{\text{unified}}$ and $\sigma(\cdot)$ is the activation function(e.g., ReLU).

\subsection{Classification Head and Loss Function}
\subsubsection{Classification Head.} Let $Z \in \mathbb{R}^{(2KN) \times d}$ denote the node representations output by the final unified graph convolution layer, where $i$ is the hidden dimension. To obtain graph-level classification features, we first split the multi-band representations as \( Z = [Z^{(1)}; \ldots; Z^{(2K)}] \) to obtain the embedding of each band, where \( Z^{(k)} \in \mathbb{R}^{N \times d} \) for \( k = 1, \ldots, 2K \). For each sub-band representation \( Z^{(k)} \), we apply mean readout to obtain graph-level embeddings and the final prediction is computed by concatenating all sub-band embeddings and feeding them into an MLP:
\begin{equation}
h_k = \frac{1}{N}\sum_{i=1}^{N} Z_i^{(k)} \in \mathbb{R}^d,\quad
\hat{y} = \operatorname{MLP}\Bigl([h_1 \parallel \ldots \parallel h_{2K}]\Bigr) \in \mathbb{R}^c,
\tag{8}
\end{equation}
 where $\Vert$ denotes concatenation and $c$ is the number of classes.
\subsubsection{Loss Function.} 
In order to make model optimization easier to converge, we utilize three loss functions to guide the end-to-end training: (1) A commonly-used cross-entropy loss \( \mathcal{L}_{ce} \) for graph classification; (2)  a band diversity loss \( \mathcal{L}_{div} \) to encourage distinct frequency bands to capture complementary patterns; (3) a sparsity loss \( \mathcal{L}_{sparse} \) for cross-band matrices that highlights only the most salient cross-band connections.
\begin{equation}
\mathcal{L}_{\mathrm{div}} 
= \frac{1}{2K\,(2K - 1)}
  \sum_{i=1}^{2K}
  \sum_{\substack{j=1 \\ j \neq i}}^{2K}
  \cos\bigl(h_i, h_j\bigr),
\quad
\mathcal{L}_{\mathrm{sparse}}
= \left\|\mathbf{S}^{(s)} \mathbf{M}^{(m)\top}\right\|_1,
\tag{9}
\end{equation}
where  $\cos(\cdot)$ computes cosine similarity. \( \mathcal{L}_{div} \) operates on graph-level embeddings  $\{h_k\}_{k=1}^{2K}$ and  \(\|\cdot\|_1\) is \(L_1\) penalty. The total loss is computed by:
\begin{equation}
\mathcal{L}_{\text{total}} 
= \mathcal{L}_{\text{cls}} 
+ \lambda_1 \mathcal{L}_{\text{div}} 
+ \lambda_2 \mathcal{L}_{\text{sparse}},
\tag{10}
\end{equation}
where $\lambda_1$ and $\lambda_2$ are trade-off hyperparameters for balancing different losses.

\section{Experiments and Results}
\subsection{Experiment Settings}
\subsubsection{Datasets.} We use two brain network datasets constructed by Xu \textit{et al.}~\cite{xu2023data} for evaluation, which are ADNI~\cite{craddock2013neuro} for Alzheimer’s Disease (AD) and ABIDE~\cite{dadi2019benchmarking} for Autism (ASD). ADNI is categorized into four classes based on the progression of cognitive impairment: cognitive normal (CN), significant memory concern (SMC), mild cognitive impairment (MCI) and Alzheimer's disease (AD). The dataset comprises 914 CN, 73 SMC, 264 MCI, and 65 AD samples. The time series for each sample of AD has a length of 197.  For ABIDE dataset, we focus on the binary classification task of TC vs. ASD, with 61 (44.2\%) samples from patients with ASD. We truncate the first 300 time points for samples of ABIDE. For brain region parcellation, we employed the AAL116 atlas~\cite{tzourio2002automated} on both datasets.
\subsubsection{Implementation details.} 
All experiments were conducted on a Linux server with a NVIDIA GeForce RTX 4090 with 24GB memory. The whole network is trained in an end-to-end manner using the Adam optimizer with an initial learning rate of $1 \times 10^{-3}$, a weight decay of $1 \times 10^{-4}$ and a batch size of 32. We use early stopping based on AUROC metric, terminating training if AUROC does not improve for 15 consecutive epochs. The data is split to 8:1:1 for training, validation, and testing with 10-fold cross-validation. In our experiments, we found that setting the decomposition level $K=2$ in the Adaptive Cascade Decomposer acheived the best performance on both the ADNI and ABIDE datasets.
\subsection{Results}
\subsubsection{Baseline Models.} We carefully choose eleven well-acknowledged neural network models as our baseline methods, including : (1) General-Purpose GNNs: GCN~\cite{kipf2017semi}, GAT~\cite{velivckovic2017graph} and GraphSAGE~\cite{hamilton2017inductive}. (2) Advanced Brain Connectivity Networks: BrainNetCNN~\cite{kawahara2017brainnetcnn}, BrainGNN~\cite{li2021braingnn}, PRGNN~\cite{li2020pooling}
and Contrasformer~\cite{xu2024contrasformer}. (3) Frequency Domain Methods: Hu et al.~\cite{hu2021multi}, MFHC~\cite{zhang2017constructing} and the work of Tewarie et al.~\cite{tewarie2016integrating}. Although these approaches exploit multi-band representations, they generally lack a unified mechanism for adaptive cross-band connectivity learning. (4) Time Series Domain Methods: We adapt Leddam~\cite{yu2024revitalizing} and Autoformer~\cite{wu2021autoformer}, originally for long-term series forecasting, due to their learnable decomposition strategies. We report the classification Accuracy and AUROC over 10-fold cross-validation in Table~\ref{tab1}.

\begin{table}[!t]
\centering
\caption{Results over 10-fold-CV (Average ± Standard Deviation). The best result is highlighted in \textbf{bold} while the second-best result is in {\ul underline}.}\label{tab1}
\label{tab1}
\vspace{5pt} 
\begin{tabular*}{\textwidth}{l @{\extracolsep{\fill}} ccccc}
\hline
\multirow{2}{*}{\textbf{Methods}} & \multirow{2}{*}{\textbf{Year}} & \multicolumn{2}{c}{\textbf{ADNI}}                               & \multicolumn{2}{c}{\textbf{ABIDE}}                              \\ \cline{3-6} 
                                  &                                & \multicolumn{1}{l}{\textbf{Accuracy(\%)}} & \textbf{AUROC(\%)}      & \multicolumn{1}{l}{\textbf{Accuracy(\%)}} & \textbf{AUROC(\%)}      \\ \hline
GCN~\cite{kipf2017semi}             & 2017                           & 62.05±4.71       & 63.41±3.25       & 67.35±4.16       & 67.93±3.19       \\
GraphSAGE~\cite{hamilton2017inductive}       & 2017                           & 69.55±4.86       & 71.79±3.32       & 69.92±4.12       & 71.28±2.49       \\
GAT~\cite{velivckovic2017graph}             & 2018                           & 64.12±2.18       & 66.68±4.23       & 63.64±3.73       & 62.82±4.54       \\ \hline
BrainNetCNN~\cite{kawahara2017brainnetcnn}     & 2017                           & 73.27±4.59       & 72.46±3.66       & 72.71±2.32       & 73.37±1.22       \\
PRGNN~\cite{li2020pooling}           & 2020                           & 62.63±2.29       & 60.32±1.71       & 67.34±2.80       & 68.52±1.85       \\
BrainGNN~\cite{li2021braingnn}        & 2021                           & {\ul 74.31±1.62} & 69.73±2.39       & {\ul 74.92±1.68} & {\ul 74.20±2.12} \\
Contrasformer~\cite{xu2024contrasformer}   & 2024                           & 73.32±4.04       & 69.26±1.03       & 73.06±1.35       & 73.53±2.86       \\ \hline
Tewarie et al.~\cite{tewarie2016integrating} & 2016                           & 74.18±2.34       & {\ul 72.58±1.52} & 66.32±4.14       & 67.58±4.45       \\
MFHC~\cite{zhang2017constructing}            & 2017                           & 68.26±1.64       & 71.94±2.41       & 71.27±2.49       & 71.05±2.73       \\
Hu et al.~\cite{hu2021multi}      & 2021                           & 67.24±3.85       & 72.31±2.96       & 66.28±1.27       & 67.06±2.58       \\ \hline
Autoformer~\cite{wu2021autoformer}      & 2021                           & 65.29±3.24       & 63.85±2.16       & 68.49±4.78       & 67.35±3.52       \\
Leddam~\cite{yu2024revitalizing}          & 2024                           & 70.35±3.77       & 65.31±3.41       & 70.22±1.25       & 69.42±1.75       \\ \hline
Ada-FCN (Ours)                    & -                              & \textbf{79.68±2.65}      & \textbf{75.30±1.24} & \textbf{77.89±1.52}      & \textbf{77.62±1.68} \\ \hline
\end{tabular*}
\end{table}

\subsubsection{Ablation Studies.} 

To demonstrate the effectiveness of our proposed Ada-FCN model, we conducted ablation studies focusing on key components.
Table~\ref{tab2} presents the results. DT denotes the dynamic threshold approach, where ablation experiments are performed using the first 25\% as the threshold value. On both datasets, removing the DT significantly reduces accuracy, indicating that DT is crucial for model performance. Removing $\mathcal{L}_{\text{spasity}}$
 decreases Accuracy on both datasets. This suggests that the sparsity constraint helps the model to learn more representative brain network features. Removing $\mathcal{L}_{\text{div}}$ also decreases Accuracy on both datasets. Ablation experiments show that each component of the Ada-FCN model contributes to the final performance.


\begin{table}[!t]
\centering
\caption{Ablation study on important components in Ada-FCN on ADNI and ABIDE datasets. The best result is highlighted in \textbf{bold}.}
\label{tab2}
\vspace{5pt} 
\begin{tabular*}{0.8\textwidth}{c @{\extracolsep{\fill}} cccc}
\hline
\multicolumn{3}{c}{} & \textbf{ADNI} & \textbf{ABIDE} \\ 
\cline{4-5} 
DT                        & $\mathcal{L}_{sparse}$    & $\mathcal{L}_{div}$       & \textbf{Accuracy (\%)} & \textbf{Accuracy (\%)} \\ \hline
\checkmark & \checkmark & \checkmark & \textbf{79.68±1.65}   & \textbf{77.89±1.52}   \\
\checkmark & \checkmark &                           & 77.35±1.22            & 76.53±1.95            \\
\checkmark &                           & \checkmark & 76.14±1.66            & 73.29±1.61            \\
                          & \checkmark & \checkmark & 79.33±1.29            & 76.26±1.14            \\ \hline
\end{tabular*}
\end{table}

\begin{figure}[!t]
\includegraphics[width=\textwidth]{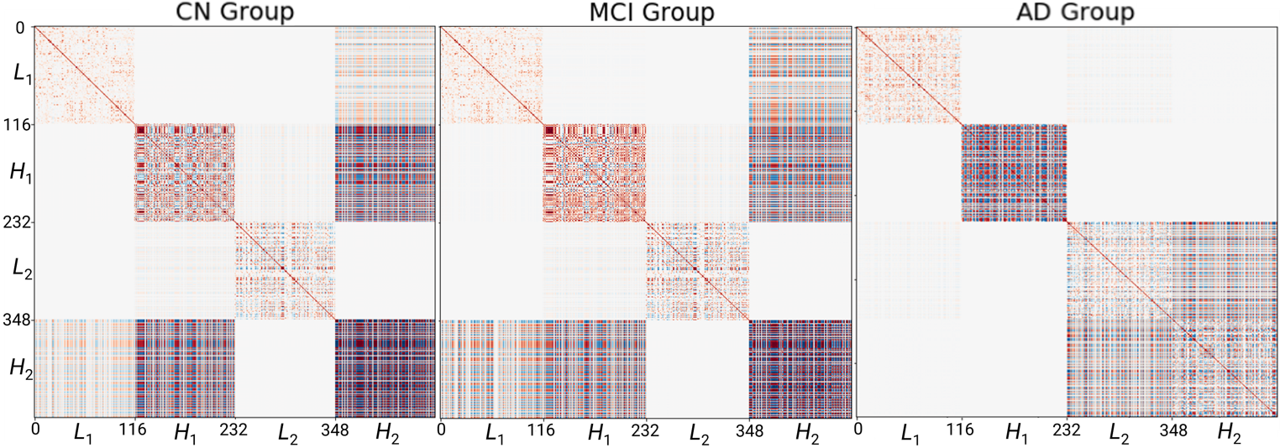}
\caption{Distinct frequency-coupled connectivity patterns revealed by group-averaged $A_{\text{unified}}$ matrices for CN, MCI, and AD.} \label{pic6}
\end{figure}

\subsubsection{Interpretability.} 
We visualize $A_{\text{unified}}$, which integrated intra- and cross-band connectivity across four sub-bands, revealed distinct group differences. As shown in Fig.~\ref{pic6}. 
The AD group exhibited significantly stronger $H_1$ intra-band connectivity than CN and MCI, suggesting a potential role for altered high-frequency oscillations in AD. While some cross-band connections (e.g., $L_1$-$H_1$, $L_2$-$H_2$) were similar across groups, AD generally showed weaker cross-frequency interactions.  While the $L_1$-$L_2$ cross-band connectivity was found to be uniformly weak across all groups, the MCI group uniquely demonstrated a prominent enhancement of intra-band connectivity within the $H_1$ frequency band. Furthermore, MCI often presented connectivity patterns intermediate between AD and CN, consistent with its potential as a prodromal AD stage. 

\section{Conclusion}
In this paper, we present Ada-FCN, a novel framework for neurological disorder classification using fMRI data, designed to overcome the limitations of prior methods that disregard inherent multi-frequency characteristics of fMRI. Ada-FCN adaptively extracts task-relevant frequency sub-bands, moving beyond the constraints of pre-defined frequency ranges, and constructs a unified brain functional network by capturing both intra-band and complex cross-band interactions, leading to refined node representations for enhanced classification. Future work will focus on examining the interpretability of the learned frequency-specific features.

\medskip\noindent\textbf{Acknowledgments.} 
\small 
This work was partially supported by RGC Collaborative Research Fund (No. C5055-24G), the Start-up Fund of The Hong Kong Polytechnic University (No. P0045999), the Seed Fund of the Research Institute for Smart Ageing (No. P0050946), and Tsinghua-PolyU Joint Research Initiative Fund (No. P0056509). The authors also acknowledge the use of AI tools for assistance in grammar enhancement and spelling checks during the preparation of this manuscript.

\begin{credits}
\subsubsection{\discintname}
\small 
The authors have no competing interests to declare that are relevant to the content of this article.
\end{credits}

\bibliographystyle{splncs04}
\bibliography{Main}

@String(ICLR = {Int. Conf. Learn. Represent.})

@String(ICLR  = {ICLR})

@article{li2021braingnn,
  title={{BrainGNN}: Interpretable brain graph neural network for {fMRI} analysis},
  author={Li, Xiaoxiao and Zhou, Yuan and Dvornek, Nicha and Zhang, Muhan and Gao, Siyuan and Zhuang, Juntang and Scheinost, Dustin and Staib, Lawrence H and Ventola, Pamela and Duncan, James S},
  journal={Medical Image Analysis},
  volume={74},
  pages={102233},
  year={2021},
  publisher={Elsevier}
}

@article{kawahara2017brainnetcnn,
  title={{BrainNetCNN}: Convolutional neural networks for brain networks; towards predicting neurodevelopment},
  author={Kawahara, Jeremy and Brown, Colin J and Miller, Steven P and Booth, Brian G and Chau, Vann and Grunau, Ruth E and Zwicker, Jill G and Hamarneh, Ghassan},
  journal={NeuroImage},
  volume={146},
  pages={1038--1049},
  year={2017},
  publisher={Elsevier}
}

@misc{yu2025linoadvancingrecursiveresidual,
      title={LiNo: Advancing Recursive Residual Decomposition of Linear and Nonlinear Patterns for Robust Time Series Forecasting}, 
      author={Guoqi Yu and Yaoming Li and Xiaoyu Guo and Dayu Wang and Zirui Liu and Shujun Wang and Tong Yang},
      year={2025},
      eprint={2410.17159},
      archivePrefix={arXiv},
      primaryClass={cs.LG},
      url={https://arxiv.org/abs/2410.17159}, 
}

@misc{yu2025refocusreinforcingmidfrequencykeyfrequency,
      title={ReFocus: Reinforcing Mid-Frequency and Key-Frequency Modeling for Multivariate Time Series Forecasting}, 
      author={Guoqi Yu and Yaoming Li and Juncheng Wang and Xiaoyu Guo and Angelica I. Aviles-Rivero and Tong Yang and Shujun Wang},
      year={2025},
      eprint={2502.16890},
      archivePrefix={arXiv},
      primaryClass={cs.LG},
      url={https://arxiv.org/abs/2502.16890}, 
}

@inproceedings{xu2023data,
  title={Data-Driven Network Neuroscience: On Data Collection and Benchmark},
  author={Xu, Jiaxing and Yang, Yunhan and Huang, David Tse Jung and Gururajapathy, Sophi Shilpa and Ke, Yiping and Qiao, Miao and Wang, Alan and Kumar, Haribalan and McGeown, Josh and Kwon, Eryn},
  booktitle={Thirty-seventh Conference on Neural Information Processing Systems Datasets and Benchmarks Track},
  year={2023}
}

@inproceedings{kipf2017semi,
  title={Semi-Supervised Classification with Graph Convolutional Networks},
  author={Kipf, Thomas N. and Welling, Max},
  booktitle={International Conference on Learning Representations (ICLR)},
  year={2017}
}

@article{hamilton2017inductive,
  title={Inductive representation learning on large graphs},
  author={Hamilton, Will and Ying, Zhitao and Leskovec, Jure},
  journal={Advances in neural information processing systems},
  volume={30},
  year={2017}
}

@article{velivckovic2017graph,
  title={Graph attention networks},
  author={Veli{\v{c}}kovi{\'c}, Petar and Cucurull, Guillem and Casanova, Arantxa and Romero, Adriana and Lio, Pietro and Bengio, Yoshua},
  journal={arXiv preprint arXiv:1710.10903},
  year={2017}
}

@inproceedings{li2020pooling,
  title={Pooling regularized graph neural network for fmri biomarker analysis},
  author={Li, Xiaoxiao and Zhou, Yuan and Dvornek, Nicha C and Zhang, Muhan and Zhuang, Juntang and Ventola, Pamela and Duncan, James S},
  booktitle={Medical Image Computing and Computer Assisted Intervention--MICCAI 2020: 23rd International Conference, Lima, Peru, October 4--8, 2020, Proceedings, Part VII 23},
  pages={625--635},
  year={2020},
  organization={Springer}
}

@article{dadi2019benchmarking,
  title={Benchmarking functional connectome-based predictive models for resting-state fMRI},
  author={Dadi, Kamalaker and Rahim, Mehdi and Abraham, Alexandre and Chyzhyk, Darya and Milham, Michael and Thirion, Bertrand and Varoquaux, Ga{\"e}l and Alzheimer's Disease Neuroimaging Initiative and others},
  journal={NeuroImage},
  volume={192},
  pages={115--134},
  year={2019},
  publisher={Elsevier}
}

@article{craddock2013neuro,
  title={The neuro bureau preprocessing initiative: open sharing of preprocessed neuroimaging data and derivatives},
  author={Craddock, Cameron and Benhajali, Yassine and Chu, Carlton and Chouinard, Francois and Evans, Alan and Jakab, Andr{\'a}s and Khundrakpam, Budhachandra Singh and Lewis, John David and Li, Qingyang and Milham, Michael and others},
  journal={Frontiers in Neuroinformatics},
  volume={7},
  pages={27},
  year={2013}
}

@article{kan2022brain,
  title={Brain network transformer},
  author={Kan, Xuan and Dai, Wei and Cui, Hejie and Zhang, Zilong and Guo, Ying and Yang, Carl},
  journal={Advances in Neural Information Processing Systems},
  volume={35},
  pages={25586--25599},
  year={2022}
}

@article{ying2021transformers,
  title={Do transformers really perform badly for graph representation?},
  author={Ying, Chengxuan and Cai, Tianle and Luo, Shengjie and Zheng, Shuxin and Ke, Guolin and He, Di and Shen, Yanming and Liu, Tie-Yan},
  journal={Advances in neural information processing systems},
  volume={34},
  pages={28877--28888},
  year={2021}
}

@inproceedings{cui2022interpretable,
  title={Interpretable graph neural networks for connectome-based brain disorder analysis},
  author={Cui, Hejie and Dai, Wei and Zhu, Yanqiao and Li, Xiaoxiao and He, Lifang and Yang, Carl},
  booktitle={International Conference on Medical Image Computing and Computer-Assisted Intervention},
  pages={375--385},
  year={2022},
  organization={Springer}
}

@article{wu2021autoformer,
  title={Autoformer: Decomposition transformers with auto-correlation for long-term series forecasting},
  author={Wu, Haixu and Xu, Jiehui and Wang, Jianmin and Long, Mingsheng},
  journal={Advances in neural information processing systems},
  volume={34},
  pages={22419--22430},
  year={2021}
}

@article{yu2024revitalizing,
  title={Revitalizing multivariate time series forecasting: Learnable decomposition with inter-series dependencies and intra-series variations modeling},
  author={Yu, Guoqi and Zou, Jing and Hu, Xiaowei and Aviles-Rivero, Angelica I and Qin, Jing and Wang, Shujun},
  journal={arXiv preprint arXiv:2402.12694
        
        
        
        },
  year={2024}
}

@article{tewarie2016integrating,
  title={Integrating cross-frequency and within band functional networks in resting-state MEG: A multi-layer network approach},
  author={Tewarie, Prejaas and Hillebrand, Arjan and van Dijk, Bob W and Stam, Cornelis J and O'Neill, George C and Van Mieghem, Piet and Meier, Jil M and Woolrich, Mark W and Morris, Peter G and Brookes, Matthew J},
  journal={Neuroimage},
  volume={142},
  pages={324--336},
  year={2016},
  publisher={Elsevier}
}

@inproceedings{zhang2017constructing,
  title={Constructing multi-frequency high-order functional connectivity network for diagnosis of mild cognitive impairment},
  author={Zhang, Yu and Zhang, Han and Chen, Xiaobo and Shen, Dinggang},
  booktitle={Connectomics in NeuroImaging: First International Workshop, CNI 2017, Held in Conjunction with MICCAI 2017, Quebec City, QC, Canada, September 14, 2017, Proceedings 1},
  pages={9--16},
  year={2017},
  organization={Springer}
}

@article{hu2021multi,
  title={Multi-band brain network analysis for functional neuroimaging biomarker identification},
  author={Hu, Rongyao and Peng, Ziwen and Zhu, Xiaofeng and Gan, Jiangzhang and Zhu, Yonghua and Ma, Junbo and Wu, Guorong},
  journal={IEEE transactions on medical imaging},
  volume={40},
  number={12},
  pages={3843--3855},
  year={2021},
  publisher={IEEE}
}

@article{yang2020frequency,
  title={Frequency-dependent changes in fractional amplitude of low-frequency oscillations in Alzheimer’s disease: a resting-state fMRI study},
  author={Yang, Liu and Yan, Yan and Li, Yuxia and Hu, Xiaochen and Lu, Jie and Chan, Piu and Yan, Tianyi and Han, Ying},
  journal={Brain imaging and behavior},
  volume={14},
  pages={2187--2201},
  year={2020},
  publisher={Springer}
}

@article{zuo2010oscillating,
  title={The oscillating brain: complex and reliable},
  author={Zuo, Xi-Nian and Di Martino, Adriana and Kelly, Clare and Shehzad, Zarrar E and Gee, Dylan G and Klein, Donald F and Castellanos, F Xavier and Biswal, Bharat B and Milham, Michael P},
  journal={Neuroimage},
  volume={49},
  number={2},
  pages={1432--1445},
  year={2010},
  publisher={Elsevier}
}

@article{tzourio2002automated,
  title={Automated anatomical labeling of activations in SPM using a macroscopic anatomical parcellation of the MNI MRI single-subject brain},
  author={Tzourio-Mazoyer, Nathalie and Landeau, Brigitte and Papathanassiou, Dimitri and Crivello, Fabrice and Etard, Olivier and Delcroix, Nicolas and Mazoyer, Bernard and Joliot, Marc},
  journal={Neuroimage},
  volume={15},
  number={1},
  pages={273--289},
  year={2002},
  publisher={Elsevier}
}

@article{xu2024contrastive,
  title={Contrastive Graph Pooling for Explainable Classification of Brain Networks},
  author={Xu, Jiaxing and Bian, Qingtian and Li, Xinhang and Zhang, Aihu and Ke, Yiping and Qiao, Miao and Zhang, Wei and Sim, Wei Khang Jeremy and Guly{\'a}s, Bal{\'a}zs},
  journal={IEEE Transactions on Medical Imaging},
  year={2024},
  publisher={IEEE}
}

@inproceedings{xu2024contrasformer,
  title={Contrasformer: A Brain Network Contrastive Transformer for Neurodegenerative Condition Identification},
  author={Xu, Jiaxing and He, Kai and Lan, Mengcheng and Bian, Qingtian and Li, Wei and Li, Tieying and Ke, Yiping and Qiao, Miao},
  booktitle={Proceedings of the 33rd ACM International Conference on Information and Knowledge Management},
  pages={2671--2681},
  year={2024}
}

@inproceedings{
xu2025brainood,
title={Brain{OOD}: Out-of-distribution Generalizable Brain Network Analysis},
author={Jiaxing Xu and Yongqiang Chen and Xia Dong and Mengcheng Lan and Tiancheng HUANG and Qingtian Bian and James Cheng and Yiping Ke},
booktitle={The Thirteenth International Conference on Learning Representations},
year={2025},
url={https://openreview.net/forum?id=3xqqYOKILp}
}

@article{peng2025biologically,
  title={Biologically Plausible Brain Graph Transformer},
  author={Peng, Ciyuan and Huang, Yuelong and Dong, Qichao and Yu, Shuo and Xia, Feng and Zhang, Chengqi and Jin, Yaochu},
  journal={arXiv preprint arXiv:2502.08958},
  year={2025}
}

\end{document}